# Spatial-Temporal Deep Embedding for Vehicle Trajectory Reconstruction from High-Angle Video

Tianya T. Zhang, *Ph.D.*, Peter J. Jin, *Ph.D.*, Han Zhou, Benedetto Piccoli, *Ph.D.*

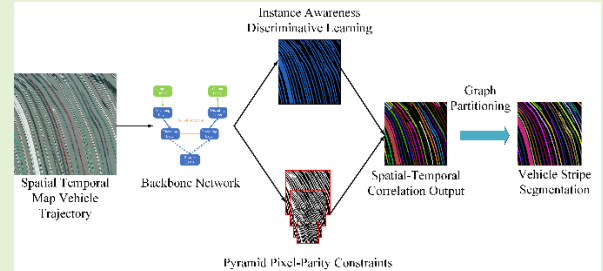

*Abstract*— **Spatial-temporal Map (STMap)-based methods have shown great potential to process high-angle videos for vehicle trajectory reconstruction, which can meet the needs of various data-driven modeling and imitation learning applications. In this paper, we developed Spatial-Temporal Deep Embedding (STDE) model that imposes parity constraints at both pixel and instance levels to generate instance-aware embeddings for vehicle stripe segmentation on STMap. At pixel level, each pixel was encoded with its 8-neighbor pixels at different ranges and this encoding is subsequently used to guide a neural network to learn the embedding mechanism. At the instance level, a discriminative loss function is designed to pull pixels belonging to the same instance closer and separate mean value of different instances far apart in the embedding space. The output of the spatial-temporal affinity is then optimized by the mutex-watershed algorithm to obtain final clustering results. Based on segmentation metrics, our model outperformed five other baselines that have been used for STMap processing and shows robustness under the influence of shadows, static noises, and overlapping. The designed model is applied to process all public NGSIM US-101 videos to generate complete vehicle trajectories, indicating a good scalability and adaptability. Last but not least, strengths of scanline method with STDE and future directions were discussed. Code, STMap dataset and video trajectory are made publicly available in the online repository. Github Link: shorturl.at/jkIT0.**

## I. Introduction

DETAILED high-quality vehicle trajectory data are of great significance for developing driving behavior models, traffic state estimation, and self-driving control algorithms. With realistic microscopic longitudinal and lateral maneuverers, valuable vehicle trajectory data lay the foundation for many scientific discoveries. There are three types of commonly used trajectory datasets. The first type is NGSIM-like trajectory [1] collected using overhead cameras that monitor a selected roadway segment from Drones or high-rising infrastructure, containing information about vehicle movements and infrastructure. The second type is collected from mobile platforms that are carried around by self-driving vehicles (e.g., KITTI [2], Waymo data [3], Argoverse [4], and ApolloScape [5]). The third type of dataset is probe vehicle trajectory like the Strategic Highway Research Program 2 Naturalistic Driving Study (SHRP2-NDS [6]) that is designed to analyze the driving behaviors of long-term participants.

The conventional tracking-by-detection video-based vehicle trajectory extraction method consists of two major steps. The first step applies an object detector for localization and classification. During the second step, data association is formulated between confirmed tracklets and newly received bounding boxes. For the object detection module, by virtue of similarities among different classes of vehicles (e.g., Car, Pickup Truck, and Mail Truck) in most traffic scenes, some vehicles, especially small trucks, could cause intra-class detections of both car and truck. For large freight trucks, the neural network model often generates double detections; in other words, the first bounding box covers only the engine, and the second bounding box covers trailer and/or engine parts. Incorporating high-performance deep learning models for multi-object tracking for the tracking module will introduce significant overheads to the systems when the state-of-the-art detections already demand high computation resources. The tracking modules often produce many broken trajectories due to the loss of tracklets, which require significant effort for post-

This work is funded by the C2SMART Project (USDOT Award #: 69A3551747124 ); New Brunswick Innovation Hub Smart Mobility Testing Ground (SMTG) Contract #: 21-60168.

T. Zhang is a Postdoc Associate at Rutgers University, NJ 08854 USA (E-mail: tz140@scarletmail.rutgers.edu).
P. J. Jin, is an associate professor at Civil and Environmental Engineering Department in Rutgers University, Piscataway, NJ 08854 USA (E-mail: peter.j.jin@rutgers.edu)
Han Zhou researcher on intelligent supply chain system design at JD.com. (E-mail: vincezhou@126.com).
Benedetto Piccoli is Joseph and Loretta Lopez Chair Professor of Mathematics Department of Mathematical Sciences. (E-mail: piccoli@camden.rutgers.edu).

processing and stitching. Most post-processing steps rely on extrapolation, that often becomes another error source.

Some recent studies [7,8] have shown that using the longitudinal scanline method to extract vehicle trajectory is reliable and effective, a one-stage vehicle trajectory reconstruction method. In contrast to the two-stage methods, the scanline method is a segmentation-is-tracking solution that converts the vehicle detection and tracking task into an image segmentation task. The first step of the scanline method is to define the scanline location for each lane and generate Spatial-Temporal Map (STMap) that accumulates vehicles' pixel movements over continuous video frames. The next step is to segment moving vehicle stripes on STMap and acquire vehicle boundaries. In previous research, the STMap segmentation model is based on semantic segmentation that classifies each pixel into either backgrounds or moving targets. The semantic segmentation is problematic when different vehicle stripes are overlayed on each other, resulting in missed detection due to the imperfect partition. This manuscript presents a new instance segmentation approach tailored to STMap vehicle trajectory reconstruction.

This proposal-free instance segmentation generates instance-aware embeddings to pull pixels within the same instance closer and, at the same time, push different instances apart. Each instance corresponds to an individual vehicle stripe on the STMap. To be more specific, we use parity constraints on both pixel level and instance level to supervise the deep neural network to learn the embedding spaces in a proposal-free manner. To achieve this goal, we have developed a specialized loss function for training and built different backbone networks that concatenate multi-resolution feature maps.

## II. RELATED WORK

### A. Vehicle Detection and Tracking

Object Detection: Object detection has been one of the fundamental problems in computer vision for the analysis of object behaviors to support a variety of tasks such as movement counting, travel behavior analysis, traffic state estimation, etc. In the early period, background subtraction was the most used method and feature-based moving target detection. These approaches are vulnerable to sudden changes in illumination, sleeping objects, or camouflage problems. Traditional machine learning approaches such as Support Vector Machines (SVM), Random Forests, or Eigen vehicles rely on hand-crafted features. Over recent years, deep learning models have become the most important solution to tackle the object detection problem. Convolutional Neural Networks using a data-driven approach can automatically extract features in an end-to-end fashion. The effectiveness of deep neural networks (DNN) has been demonstrated on large public benchmark datasets, including the MS-COCO Detection Challenge [9] and ImageNet Large Scale Visual Recognition Challenge [10]. One-stage and two-stage detections are two common categories for deep learning models. The two-stage framework contains a region proposal network to generate proposals for possible positions of objects. The proposals were then refined into bounding boxes for accurate localization. Some well-known two-stage detectors are Faster R-CNN [11] and Mask R-CNN [12]. The one-stage framework employs a single feed-forward fully convolutional network that outputs bounding box detection and classification in one pass, such as YOLO v1 ~ v5 [13-15] and SSD [16]. Lightweight models, such as MobileNet-SSD, YOLOv4-Tiny, NAS-FPNLite [17], and EfficientDet [18], were developed to reduce computational complexity while preserving the accuracy on IoT edge devices. These models have successfully balanced the speed and accuracy of high-end machines and are suitable for widespread industrial scalability.

Multi-Object Tracking: Multi-Object Tracking (MOT) is a vital component for many applications in computer vision, which is considered a post-processing task of detection outputs. The typical motif of MOT is the tracking-by-detection paradigm, where objects of interest are linked between consecutive frames while maintaining spatial-temporal consistency. Regarding MOT approaches in traffic videos, the tracking module can run either offline or online. Offline or batch processing techniques are formed as the global optimization algorithm using a batch of consecutive frames from the video or camera input. On the other hand, the Online MOT [19] uses only the current and previous frames and encoded state space for long-term dependencies of movement. This further leads to a data association between the object being tracked and the newly received detection of the MOT task. The online methods also require fast inference to perform feature association between tracklets and new detections. Commonly used methods include Kalman Filter-based [20,21], Joint probabilistic data association (JPDA) [22], Recurrent Neural Networks [23], attention and transformer-based [24,25] and Siamese Networks [26].

### B. Spatial Temporal Map Based Method

The scanline method was first introduced for motion estimation on a mobile platform to solve the structure of motion problems [27,28]. Later, this method was adopted for pedestrian detection and tracking [29,30]. Due to the efficiency of the scanline method, this method has evolved from a collection of simple traffic counting and headways [31,32] to signal phase analysis [33,34] and extraction of vehicle detection from roadside video [35-37]. In the paper [38], a video analytic system for CCTV traffic cameras was established based on the longitudinal scanline method for the NJ511 traffic camera network. The author also analyzes the cost of cloud infrastructure for potential large-scale deployment. A recent study [39] uses the spatial-temporal scanline method to evaluate image processing-based vehicle trajectories. The spatial-temporal map method provides a means to quickly assess the quality of vehicle trajectory data against the actual vehicle positions in the original video.

The previous studies are developed using traditional image processing methods, relying on edge detection, Hough Transformation, color/intensity-based classification, and motion-based analysis. In the latest research, machine learning and deep learning approaches have been applied to reconstruct vehicle trajectories from the spatial-temporal map [40, 41]. The paper [40] uses a Dynamic Mode Decomposition method to decompose the STMap into the low-rank background and sparse foreground components. The preprocessed STMaps are then used to train a deep neural network named ResUNet+, which has outperformed several high-impact image segment models. Another machine learning approach [41] incorporated two attention modules in the U-shaped structure on both spatial and channel levels. The initial backbone is replaced with Inception blocks to take full advantage of the flexible design of the Inception network. The Dual Attention Inception UNet (DAIU) model has surpassed the state-of-the-art tracking-by-detection two-stage video-based traffic solution under various conditions on different weather, illumination, and road geometry.

However, those pixel-classification background modeling or semantic segmentation-based approaches cannot deal with overlapped vehicle strands on STMap. To address the limitations of previous STMap-based trajectory detection methods, this study developed a proposal-free instance segmentation model that can distinguish individual vehicle strands and is robust to different noises.

## III. METHODOLOGY

### A. Spatial Temporal Map Generation

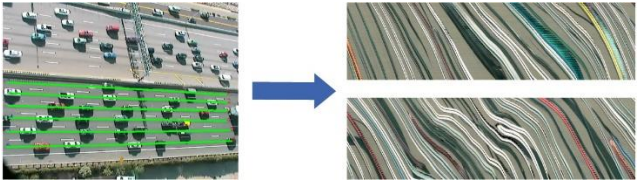

Fig. 1. Scanline Definition and Spatial-Temporal Map Generation

The scanline method is drawn on the center line of each lane in the traveling direction from the video frame, as shown (in Figure 1). The efficient Bresenham's line drawing algorithm is used to obtain each pixel coordinate on the scanline. The pixel coordinates are saved in the vector format ordered from origin to destination. Then all pixels' RGB values are stacked over continuous frames to create the Spatial-Temporal Map. Every vehicle traveling on the scanline will leave its traces shown on the STMap. Some casted shadows and lane markers are recorded as well. Our goal of vehicle trajectory detection becomes an image segmentation task to identify each vehicle strand from STMap.

Unlike commonly used image segmentation datasets such as the COCO dataset, ImageNet, or the Cityscapes dataset [42], objects on STMap share very similar textures and shapes, and the number of vehicle stripes depends on the traffic density in the camera scene. Compared to the popular biology segmentation datasets (e.g., Computer Vision Problems in Plant Phenotyping (CVPPP) dataset [43]), STMaps contain more noises (e.g., shadows and lane markers) and quality of the image could degrade due to illumination changes. In contrast to the static image, columns on STMap have temporal sequence meaning. All vehicle strands stretch from top-left to bottom-right with a similar appearance, and the scales among different vehicles are alike. The roadway background pixels on STMaps are stable and uniform, with few varieties. Those semantic peculiarities call for a different and new segmentation approach for vehicle trajectory extraction.

Most image segmentation methods are designed for datasets without temporal information. Namely each instance is a static object. However, as to the STMap, image columns represent the temporal progression of pixels' values that preserve the vehicle's movements at each timestamp. To fully capture the temporal correlation in STMap, we proposed a new pixel-level correlation learning module with spatiotemporal deep embedding (STDE). In addition, we applied an instance-level differentiation loss function to ensure the same vehicle stripe should be together in high dimensional embedding space, while different vehicle instances should be far apart. The design of this neural network has incorporated merits from several state-of-the-art proposal-free instance embedding and segmentation models [44-47].

### B. Pixel-Level Spatiotemporal Correlation Learning

This section introduces the multi-resolution spatial-temporal correlation learning module for analyzing pixelwise relationships preserved in STMap. This problem is considered correlational learning to reason the affinity relationships between center pixels and their neighbors. However, it is impossible to calculate the affinity scores for all pixel pairs. We adopted the multiscale pyramid resolution scheme to measure both the long- and short-term information, as shown in Figure 2.

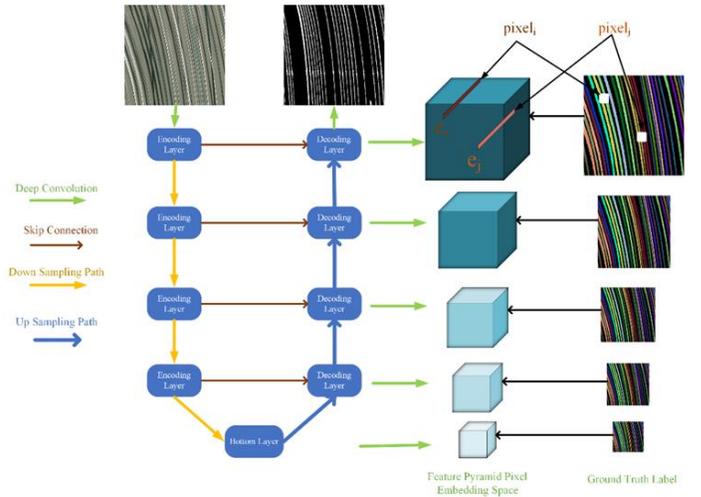

Fig. 2. Feature Pyramid Multiscale Spatial-Temporal Embedding Architecture

A STMap is defined as $S^{H*W*3}$, where H represents STMap height also representing number of pixels on scanline, and W is STMap width also representing number of video frames, 3 is R-G-B three color channel. The spatiotemporal affinity relationship of each pixel is encoded into N - dimensional vector $R = [r_1, r_2, ..., r_{N-1}, r_N]$, where N is the 8-neighbor pixels within the adjacent spatiotemporal window at different ranges (Fig. 3). And $r_i$ is the encoded value of the $i^{th}$ neighbor in its spatiotemporal adjacency list. The instance labels of all neighbors for any pixel $p$ are obtained from the ground-truth instance label, which is expressed as $l = [l_1, l_2, ..., l_{N-1}, l_N]$.

$$r_j = \begin{cases} 0, & if\ l_j = L_p \\ 1, & if\ l_j \neq L_p \end{cases}$$

Where $L_s$ is the instance label for center pixel $p$. If the neighbor pixel and the center pixel belong to the same instance, $r_j = 1$, otherwise $r_j = 0$. This neighboring correlation information will be encoded into $Y \in \mathbb{R}^{H*W*N}$ tensor. After obtaining the spatiotemporal adjacency encoding for each pixel, the next step is to learn the relationships between pixels with convolutional neural networks layers.

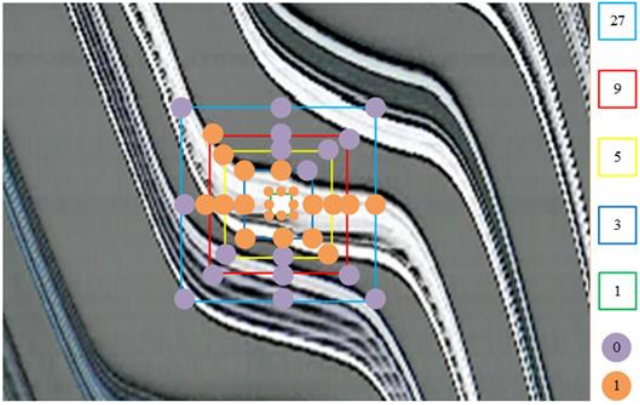

Fig. 3. Spatiotemporal Affinity Encoding for Central Pixel with 8-Neighbours at Different Ranges {1, 3, 5, 9, 27}

The input STMap $S^{H*W*3}$ will be mapped into a new space $E^{H*W*D}$ through a convolutional neural network, where $D$ represent the feature dimension for each pixel. The underlying logic for the pixel-pair spatiotemporal correlation learning is that the same instance should be closer in the feature space while distinct trajectory instances should be distant. According to several previous studies, the similarity between two embedding features is measured with cosine distance.

$$S(e_i, e_j) = \frac{1}{2}(1 + \frac{e_i^T * e_j}{\|e_i\|_2 \|e_j\|_2})$$

Where, $e_i$ and $e_j$ are embedding features for pixel $i$ and pixel $j$ on STMap. $S(e_i, e_j) = 1$ denotes those two pixels have the same embedding feature and belong to the same trajectory on STMap. $S(e_i, e_j) = 0$ denotes that those two pixels are complete opposite in the feature dimension and belong to different trajectory instance. The cosine metric is invariant to the scale of the embedding vectors, which has more flexible range than Euclidean distances. Another reason for using cosine similarity function is consistent with the definition of neighbor instance relationship $r_j$.

Based on the pairwise similarity, we can obtain another $ST$ pixel correlation tensor $\hat{Y} \in \mathbb{R}^{H*W*N}$, and then the Mean Square Error (MSE) is used to supervise our deep embedding module to learn the spatiotemporal relationship. The loss function is defined as follows:

$$\mathcal{L}_{ST} = \|\hat{Y} - Y\|_2$$
$$= \frac{1}{H*W*N} \sum_{i=1}^{N} \sum_{j=1}^{H*W} \|\hat{s}_{i,j} - s_{i,j}\|$$

As the correlations are often shown in different spatial and temporal resolution, we adopted the Feature Pyramid Network (FPN) structure to construct a top-down pathway for both high-level and low-level semantic meaning. The larger spatial-temporal window could incur huge computation cost, while a smaller spatial-temporal window would miss the long-range dependencies. Using pyramid scheme for multi-resolution information integration is a very popular technique in many image tasks. The multi-resolution spatial temporal correlation is represented by coarser to finer pyramid learning. The lower resolution in the pyramid can predict long range correlations, and the higher resolution in the pyramid can predict shorter range correlations. Also, the pyramid learning modules are conveniently to be integrated with state-of-the-art multi-level encoding-decoding architecture.

$$\mathcal{L}_d = \sum_{d \in \{\frac{1}{2}, \frac{1}{4}, \frac{1}{8}, \frac{1}{16}\}} \|\widehat{Y_d} - Y_d\|_2$$

Where $d$ represent downsampling rate at different pyramid scales. $Y_d$ represents the pixel correlation tensor at each down-sampled STMap, $\widehat{Y_d}$ is the corresponding down-sampled spatiotemporal correlation in the high-dimensional embedding space.

### C. Instance-Level Discriminative Learning

In this section, we will discuss the instance level discriminative learning module based on the premise that embeddings of the same trajectory should be at similar positions in the embedding space, while different objects should be differentiated. The previous loss function is defined by pixel-pair similarity measurement; however, the instance-level discriminative learning is defined by instance-pair distance measurement. The loss function is formulated as the sum of two terms: the concentration term $\mathcal{L}_{con}$ to guide pixels distance from the same instance and the discrepancy term $\mathcal{L}_{dis}$ to guide the distances between different instances.

In figure 2, between class average embedding vector should be distant. Pixels belonging to the same instance should be close to the average embedding vector of that class. In this Instance-Level Discriminative Learning, to reduce computational overhead, we only consider the vehicle stripes that share the same time window.

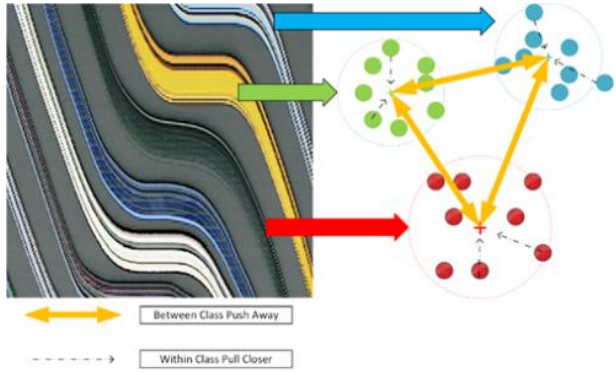

Fig. 2. Instance-Aware Discriminative Learning

The concentration of pixel embeddings within same trajectory instance is expressed as

$$\mathcal{L}_{con} = \frac{1}{\sum_{m=1}^{M} K_m} \sum_{m=1}^{K} \sum_{p=1}^{K_m} |1 - S(e_p, \mu_m)|$$

The discrepancy of pixel embeddings between different trajectory instances is expressed as:

$$\mathcal{L}_{diff} = \frac{1}{M} \sum_{m=1}^{M} \frac{1}{|N_d(m)|} \sum_{n \in N(m)} S(\mu_n, \mu_m)$$

Where, M is the total number of trajectory strands on the STMap. $\mu_m$ is the normalized mean embedding of instance $m$. $K_m$ is the total number of pixels in instance $m$. $N_\delta(m)$ are the neighbor trajectories of instance $m$, when the two trajectories share a common time window on STMap.

The discriminative loss is a balanced summation of the concentration loss and differentiation loss. And we set $\delta = \tau = 1$ following the convention in paper ([48]), as they are equally important.

$$\mathcal{L}_{dis} = \delta * \mathcal{L}_{con} + \tau * \mathcal{L}_{diff}$$

### D. Training and Inferencing

Our spatiotemporal correlation learning method is network-agnostic. Namely, the backbone image segmentation networks in the framework can be implemented with various SOTA models. In this paper, we implemented the popular and efficient ResUNet as the multi-resolution encoding-decoding architecture.

The above three loss functions are combined for the end-to-end training as:

$$\mathcal{L}_{total} = \alpha * \mathcal{L}_{ST} + \beta * \mathcal{L}_d + \gamma * \mathcal{L}_{dis}$$

where $\alpha, \beta, \gamma$ are balancing coefficients on these loss function terms.

In the inference phase, we only use the $ST$ pixel correlation tensor $\hat{Y}$ as the final classification prediction. Watershed segmentation and Mutex algorithm ([49]) are frequently used methods as post-processing to obtain individual trajectory mask from $\hat{Y}$. In addition, we aggregate small instances into large instance and eliminate too small blobs to further refine the final segmentation result.

## IV. EXPERIMENT

### A. Video Data

The high-angle Next Generation Simulation (NGSIM) video data are downloaded from the U.S. DOT Intelligent Transportation Systems (ITS) Public Data Hub, which is a real-world trajectory dataset consisting of high-resolution vehicle trajectory data, raw/processed video files and metadata to support driving behavior analysis. US-101 data consist of five main lanes and one auxiliary lane, one on-ramp, and one off-ramp from US 101 (Hollywood Freeway) located in Los Angeles, California, which is the 2100-foot-long study area divided into eight sub-areas by eight synchronized cameras. The dataset contains the trajectory data of all vehicles in the study area during the morning peak periods, 1) 7:50 a.m. to 8:05 a.m., 2) 8:05 a.m. to 8:20 a.m., and 3) 8:20 a.m. to 8:35 a.m. on June 15th, 2005. The defined scanline for all cameras were shown in Fig. 3.

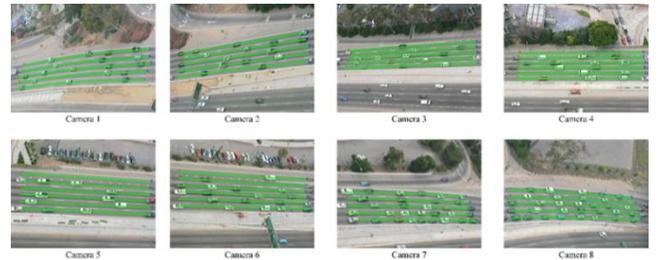

Fig. 3. Scanlines for Eight Cameras from NGSIM US 101 Video Data

The training data preparation is tackled through two approaches. In our previous studies, the author has developed image semantic segmentation models that classify pixels into vehicle stripes and background pavements. Two machine learning algorithms were applied, including Dynamic Mode Decomposition (DMD) and Res-UNet+ deep neural network, to first obtain some segmentation mask from I-80 dataset. Then, we utilize the segmentation masks to obtain instance level mask. This way, we significantly reduced the workload for training data preparation, as the data annotation is usually the most time-consuming and costly steps for any machine learning/ deep learning project.

Most of the training dataset are prepared through afore-mentioned image segmentation algorithms. Some of the training dataset that contains difficult noises, shadows and severe occlusions requires manual process to correct the pixel labels. We manually inspected the automation tool's outputs and cleaned all observable noises. The total dataset consists of 300 training data, 100 validation data, and 100 test data, following the 60%-20%-20% partition convention. The train, validate and test datasets can be found in project public repository [51].

## B. Implementation Details

We implemented two different schemes as the backbone including Attention-UNet and Residual-UNet. An adaptive learning rate optimization algorithm Adam was used with $\beta 1 = 0.9$, $\beta 2 = 0.999$. The learning rate is set as 0.0001 on an NVIDIA GeForce RTXTM 3070 GPU for 1,000 iterations. The penalty coefficients of the loss function are empirically set as $\alpha = \beta = \gamma = 1$. The number of embedding dimensions is set to 16. The image size is chosen as $512 \times 512$ and a batch size is 4. The loss function uses the Weighted MSE (Mean Square Error) on training. Symmetric best Dice coefficient (SBD) is utilized on validation dataset for model selection, which averages the intersection over union (IOU) between pairs of predicted and the ground truth labels. We found that Attention-UNet Backbone not only have smaller size of parameters (2.81 million) than Residual-UNet (4.72 million parameters), but also have lower SBD score and faster converge speed.

## V. VALIDATION

In this section, our trajectory stripe segmentation results on testing dataset under different scenarios were presented. We also examined our model outputs against five different baseline models using comprehensive segmentation metrics.

### A. Evaluation on Testing Data

In the next Fig. 4, major steps of vehicle stripe segmentation are shown, including the embedding visualization for each pixel, the raw spatial-temporal segmentation output, and post-processing using a parameter-free Mutex Watershed algorithm for cluster refinement.

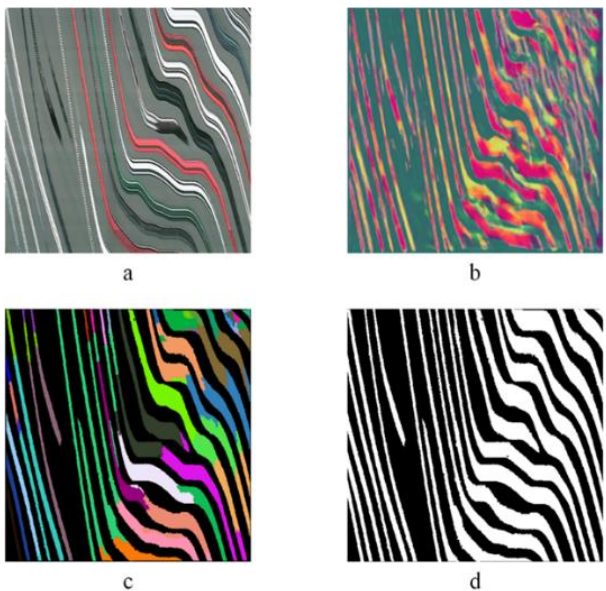

Fig. 4. Results from Spatial-Temporal Correlation Embedding. (a. STMap Input; b. Embedding Visualization; c. Spatial-Temporal Correlation Output; d. Trajectory Stripe Binary Mask Output)

To quantify the model performance, we compare the new STMap segmentation model with SOTA baselines used in previous paper [40, 41] based on comprehensive segmentation metrics. Our model shows perfect segmentation results on all five categories. Global Accuracy is calculated as the percentage of correctly classified pixels regardless of class. Mean accuracy the percentage of correctly classified pixels to the total number of pixels for each class. The IoU (also known as the Jaccard coefficient) is used to measure the agreement between two segmented image areas. Mean IoU is weighted by the number of pixels in the class. BF Score computes the harmonic mean between precision and recall value to decide whether the points on the boundary have been matched.

The spatial-temporal Deep Embedding outperformed five high-impact baseline models on all evaluation metrics as shown in Fig. 5. The FCN (Fully Convolutional Networks) is the seminal encoding-decoding network for per-pixel image segmentation, which inherited by many following deep learning models. DeepLabV3 is designed a to handle the problem of segmenting objects at multiple scales with atrous convolution in cascade. SegNet is a standard architecture consisting of encoding-decoding followed by pixel-wise classification layer. Res-UNet and Res-UNet+ are two models built on UNet architecture with residual block as backbone and extra connection on decoding layers for functionality enhancement. Res-UNet+ demonstrates excellent segmentation results on STMap based vehicle trajectory extraction, which is proposed by the same author and tested using NGSIM I-80 video dataset.

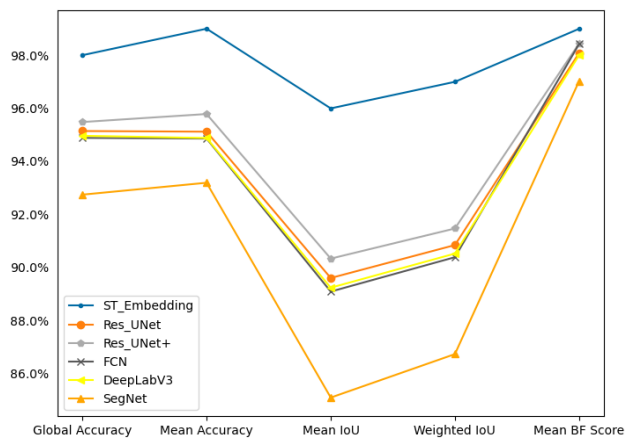

Fig. 5. Segmentation Metrics Compared to Baselines

### B. Model Performance Under Different Scenarios

In this segment, our model performance under normal conditions and challenging conditions is presented to further demonstrate the strength of proposed method. All selected images and segmentation results are from the test dataset. As shown in the testing image, the proposed STDE method generate instance level segmentation results, which address the overlapping issue caused by the semantic segmentation model that only output pixel level classification.

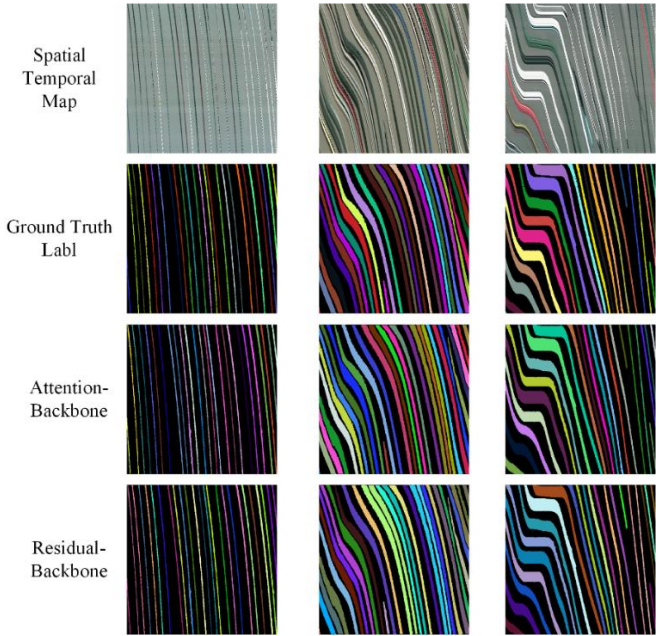

Fig. 8. Spatial Temporal Deep Embedding under Normal Condition

Under the normal condition (Figure 8), different levels of traffic density are chosen to represent the free flow traffic, traffic with slight oscillations, and traffic congestion with complete stops. In the challenging conditions, spatial-temporal map with static road markings, and two severe shadow scenarios were demonstrated. Vehicle stripe segmentation results using both Attention backbone and Residual backbone are provided. The results show that our model is robust and consistent under the influence of static noise and shadows.

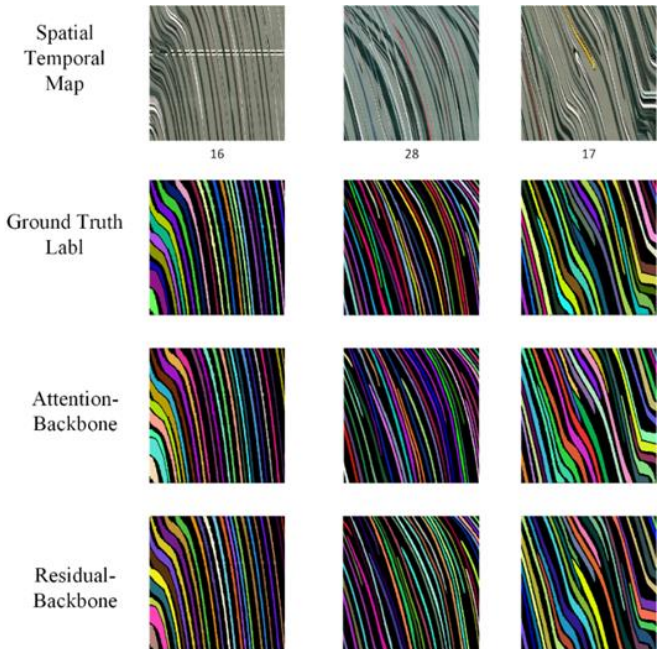

Fig. 9. Spatial Temporal Deep Embedding under Challenging Conditions (Static Noise, Shadows, and Occlusions)

One of the outstanding issues regarding STMap vehicle trajectory extraction is the shadow casted from adjacent lanes that resemble the normal vehicle stripes (Figure 9). This issue was not handled well with previous image processing-based shadow removal or semantic segmentation deep learning models based on pixel classification. However, by applying the spatial-temporal embedding to learn the correlation of pixels and the differentiation loss functions, the shadow issues were suppressed successfully in the final output.

## VI. VIDEO TRAJECTORY RECONSTRUCTION

The final step is to extract vehicle trajectories from partitioned vehicle stripes in spatial-temporal map. In OpenCV, it only needs one line function cv.findContours() to extract the boundaries of vehicle stripes. To classify whether the boundary points are front or rear bumper, the following algorithm is implemented.

Algorithm: Classify Boundary Pixels as Front or Rear Bumper
========================================
Input: Instance Segmentation Output
Outputs: Vehicle Boundary Pixel Trajectory
---------------------------------------------------------------------
**For** each instance
  Run cv.findContours() function to find contour pixels for each instance
  **For** each boundary pixel $P(x, y)$
    Count number ($N_{TopRight}$) of pixels with same label in a top-right box of height (h), width (w)
    Count number ($N_{BottomLeft}$) of pixels with same label in a bottom-left box of height (h), width (w)
    If $N_{TopRight} > N_{BottomLeft}$
      Mark $P(x, y)$ as front bumper
    Otherwise
      Mark $P(x, y)$ as rear bumper
  **End**
**End**

Compared to conventional cascaded multi-step video analytics, the great advantage of scanline method is that trajectory extraction is handled as an integrated process. The vehicle trajectory on STMap is shown by the **Error! Reference source not found.**. To avoid the self-occlusion issue caused by tilted camera angle, for camera 1 to 3, the rear bumper is extracted, as vehicles traveled away from the scene. For camera 4 to camera 8, the front bumper is extract, as the vehicles moved towards the camera location.

Finally, all video recordings from available NGSIM US-101 datasets were processed to demonstrate the generalization capability of proposed method. Here is the snapshot from one camera with congested traffic condition. A lane change scenario was especially called out when a red vehicle conducted the lane-change maneuver to hop on a faster lane (Fig. 7). The purpose of showing a lane change behavior is to indicate that by considering multiple longitudinal scanlines at the same time, we can also extract lateral driving behavior for lane change analysis. The link of all vehicle trajectory detection videos can be found in our public repository.

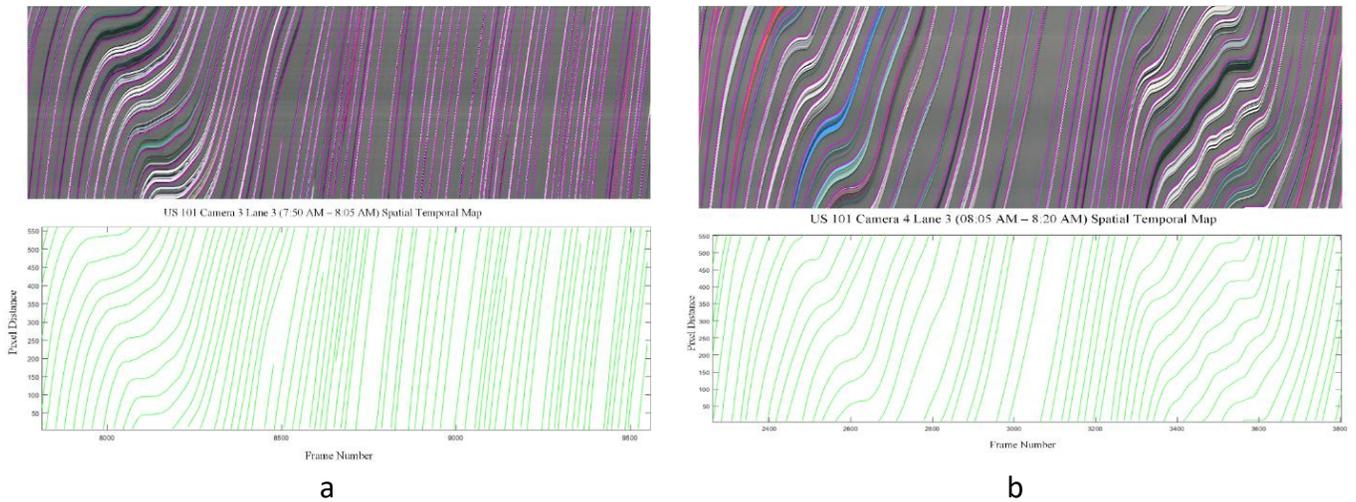

Fig. 6. a. Extracted Vehicle Rear Bumper Trajectory from STMap. b. Extracted Vehicle Front Bumper Trajectory from STMap (STMap is flipped upside down to be Consistent with Plotted Trajectory)

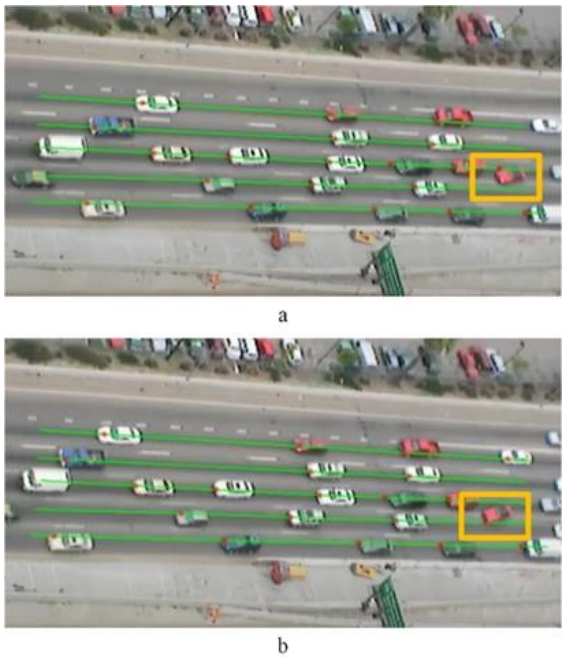

Fig. 7. Vehicle Trajectory Plotted on Videos (a. Lane Change Initiated; b. Touched on Targeted Lane)

## VII. CONCLUSION AND OUTLOOK

In this study, we developed a spatial-temporal deep embedding approach for STMap vehicle stripe segmentation considering pixel-level similarity and instance level discrimination. The feature pyramid approach is applied to address both long- and short-range dependencies. To reduce the computational needs, we only consider vehicle trajectories that share a common temporal window. Two backbones, attention-based and residual-based UNet structures, were implemented, showing that the designed module is compatible with various feature encoding networks. Based on the segmentation metrics, the proposed model surpassed several baselines and demonstrated better robustness under the influence of shadow, overlapping and static noise. After model selection and validation, we applied the trained model to process all published videos from NGSIM US-101 dataset and extracted vehicle trajectories in large scope, indicating a good scalability of the model.

The STMap offers an efficient way to validate the video trajectory dataset. Although NGSIM dataset were considered as ground truth since it has been launched, the systematic errors were investigated and reported in several studies [51-54]. The Fig. 8 shows a collision of two vehicles marked within the yellow box. Most of the trajectory errors were found during the shockwave conditions, where vehicles change their speed in response to preceding vehicle's speed change.

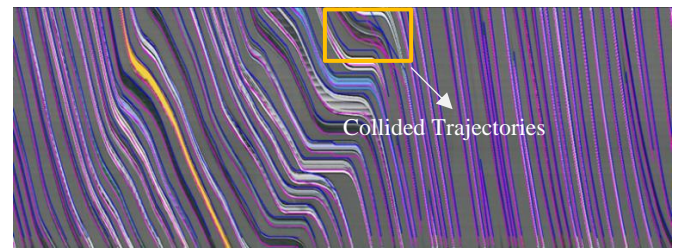

Fig. 8. NGSIM trajectory Validation Using STMap (Purple: Front Bumper; Blue: Rear Bumper)

However, current data clean approach can only address the problem with statistic smoothing approaches, none of exiting methods could efficiently identify and correct the trajectory at microscopic level. The original quality control step of NGSIM project employed very laborious process through human intervention. With the STMap method, we can exactly pinpoint the tracking error without the need of watching the video playback.

High-angle traffic video data become more accessible than ever before by virtue of improvement of computer vision software, prevalent camera devices, and drone based overhead platform. On one hand, the low-cost video-based traffic operational systems provide enormous opportunity for data acquisition. On the other hand, data-driven machine learning and imitation learning models are hungry for high-quality data input. However, the video-based trajectory data collection is prone to errors and require substantial amount of labor and expense to conduct data cleaning and post-processing. With Spatial-Temporal Deep Embedding model, the scanline method could serve as a promising method to reconstruct and fix the errors found in the legacy NGSIM dataset and removed all irrational movements due to the imperfect video processing software program.

## ACKNOWLEDGEMENT

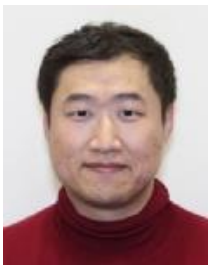

**Tianya Terry Zhang** received PhD degree in Transportation Engineering from the Department of Civil and Environmental Engineering at Rutgers University. He earned master's degrees in Transportation Engineering from Texas A&M University, and Computer Science from University of Pennsylvania.

He is a Postdoc Associate at the Department of Mathematical Sciences at Rutgers University–Camden, working on collaborative driving. He has been involved in several ITS projects related to Automated Traffic Signal Performance Measures (ATSPMs) and Smart Mobility Testing Ground. His research using computer vision and LiDAR for vehicle trajectory detection has been published in Transportation Research Part C and Transportation Research Record.

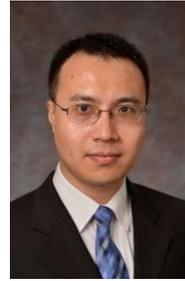

**Peter J. Jin** is an Associate Professor at Department of Civil and Environmental Engineering (CEE) at Rutgers, The State University of New Jersey. He received the BS degree in Automation at Tsinghua University, China. He received his MS and Ph.D. degrees in civil engineering from University of Wisconsin-Madison in 2007 and 2009 respectively. He worked at Center for Transportation Research, at the University of Texas at Austin as a postdoctoral fellow and research associate. He has more than 45 peer-reviewed journal publications and more than 70 conference papers. His research interests include transportation big data analytics, intelligent transportation systems, connected and automated vehicles (CAVs), and unmanned aerial vehicles (UAVs). He holds two patents in both UAVs and CAVs.

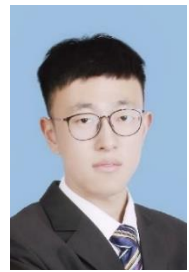

**Han Zhou** Han Zhou was born in Anhui, China, in 1995. He received the B.S. and M.S. degrees from the School of Mechanical Engineering, Beijing Institute of Technology, China, in 2019. His research interests include vehicle dynamics and electromechanical drives.

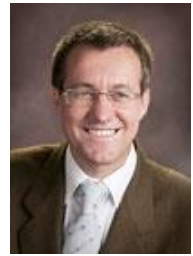

**Benedetto Piccoli** is the Joseph and Loretta Lopez Chair Professor of Mathematics in the Department of Mathematical Sciences at Rutgers University–Camden as well as the Vice Chancellor for Research of the university.